\documentclass[12pt]{article}
\usepackage{authblk}
\usepackage{geometry}
\usepackage{hyperref}
\usepackage{amsmath}
\usepackage{amssymb}
\usepackage{amsfonts}
\usepackage{enumitem}
\usepackage{graphicx}

\title{Analysis of Customer Journeys Using Prototype Detection and Counterfactual Explanations for Sequential Data}

\author{Keita Kinjo\thanks{Email: \texttt{kkinjo@kyoritsu-wu.ac.jp}; ORCID: 0000-0001-5907-3501}}
\affil{Faculty of Business, Kyoritsu Women's University\\
2-2-1, Hitotsubashi, Chiyoda-ku, Tokyo, 101-8437, Japan}

\date{}

\begin{document}

\maketitle

\begin{abstract}
Recently, the proliferation of omni-channel platforms has attracted interest in customer journeys, particularly regarding their role in developing marketing strategies. However, few efforts have been taken to quantitatively study or comprehensively analyze them owing to the sequential nature of their data and the complexity involved in analysis. In this study, we propose a novel approach comprising three steps for analyzing customer journeys. First, the distance between sequential data is defined and used to identify and visualize representative sequences. Second, the likelihood of purchase is predicted based on this distance. Third, if a sequence suggests no purchase, counterfactual sequences are recommended to increase the probability of a purchase using a proposed method, which extracts counterfactual explanations for sequential data. A survey was conducted, and the data were analyzed; the results revealed that typical sequences could be extracted, and the parts of those sequences important for purchase could be detected. We believe that the proposed approach can support improvements in various marketing activities.
\end{abstract}


\section{Introduction}

Consumer environments have become increasingly complex in recent years, with various options being offered for gaining increased product awareness, gathering information, making purchases, and sharing product reviews. Awareness can be raised via not only mass media but also web advertisements and social media. Physical stores, social media, and comparison sites are used for information gathering. Consumers perform purchases through multiple channels and omni-channels, including online e-commerce sites and physical stores \cite{ref1, ref2}. Subsequently, they commonly share product reviews on social media and online sites.

A customer journey is a method for analyzing consumer perceptions and behavior \cite{ref3}. This is a typology of a particular sequence of consumer cognition and behavior that companies use in marketing strategies \cite{ref4, ref5, ref6}. By analyzing customer journeys, companies can better understand consumers and design consumer experiences more efficiently \cite{ref7}. Specific patterns in customer journeys have been frequently extracted through qualitative research. However, the number of studies that quantitatively analyze or formalize customer journeys remains limited.

One of the few to quantitatively analyze customer journeys, Rosenbaum et al.\ (2017) investigated the frequency of consumer cognition and behavior at each stage of consumption (pre-purchase, purchase, post-purchase) \cite{ref8}. There are several studies similar to these \cite{ref5}. However, the length of the customer journey of each individual is different, and the corresponding data should be sequential in several stages. However, since sequential data are difficult to handle in tabular data analysis, there has been limited related research. For example, Herhausen et al.\ simply counted customer journey patterns \cite{ref9}.

One of the few to formalize customer journeys, Bernard and Andritsos (2017) did so (defining the data structure using trees) in a process mining framework \cite{ref10}. Other studies have employed Markov models, mixed Markov models, edit distances between sequences, or genetic algorithms to detect patterns \cite{ref11, ref12, ref13, ref14, ref15}.

Several studies have employed sequence analysis methods based on inter-sequence distance measures. For example, Topaloglu, Oztaysi, and Dogan (2024) analyzed more than 20 million rows of clickstream data collected from a real e-commerce site and proposed a four-stage framework for extracting and interpreting visitor journeys. One of these stages involved clustering based on Levenshtein distance. Their findings revealed, among other insights, that journeys characterized by exploratory behavior exhibited significantly lower conversion rates than other behavioral patterns \cite{ref16}.

In another study, Bobroske et al.\ (2020) converted sequences of actions from 10,000 patients into strings and customized the Levenshtein edit-distance algorithm to evaluate similarities between those sequences \cite{ref17}. Using this distance-based approach for clustering, they identified twelve representative patient-journey patterns.

Disadvantageously, existing research cannot suitably address the following issues. First, it is difficult to extract typical cases from sequential data that have a structured nature—such as the stages of consumption in a customer journey—which differs from general, unstructured sequences. Furthermore, in such cases, it is not only important to extract typical patterns, but also to compare them with other sequences. To enable such comparisons, the position of each pattern within the overall sequence space should be visualized and interpreted before being applied. Marketers need to identify and visualize both typical and representative customer journeys in order to gain insights into customer needs, understand their perceptions, and develop effective strategies \cite{ref18}.

In the context of marketing applications, it is necessary not only to extract typical patterns using sequence-based distance measures but also to predict outcomes such as whether a purchase will occur. For companies, whether a purchase ultimately takes place is a critical factor \cite{ref16}. This is because such predictions can help identify which types of sequences are more likely to lead to purchases and can also be used to estimate outcomes if the sequences are modified. To achieve this, the application of machine learning methods based on sequence distance is essential. However, this topic has received limited attention in existing research.

In addition, it is important to propose how a non-purchasing sequence could be improved to lead to a purchase. Such improved sequences can help uncover the factors that drive purchasing decisions. In recent years, as many machine learning models—including those using distance-based approaches—have been criticized for being difficult to interpret or functioning as ``black boxes,'' the field of explainable artificial intelligence (XAI) or interpretable machine learning has gained significant attention \cite{ref19, ref20}. Numerous techniques have been developed, including LIME and SHAP \cite{ref21, ref22}. Among these, counterfactual explanation (CE) has emerged as a promising approach \cite{ref23, ref24}. CE identifies the minimal changes required in a given instance to alter the prediction of a machine learning model. By analyzing the parts of an instance that must be changed to yield a different outcome, CE enables instance-level interpretation of relevant factors. Today, CE methods are widely used in fields such as image analysis and time series prediction. In the context of time series data, some studies have applied distance-based methods such as Dynamic Time Warping (DTW) to generate counterfactuals that are not only interpretable but also closer to typical patterns, thereby ensuring realism \cite{ref25}.

However, these methods are designed for numerical time series data. Although a few attempts have been made to apply CE to symbolic sequence data like the kind addressed in this study, they typically rely on mixed-integer linear programming (MILP), which limits their applicability to specific models \cite{ref26}. To date, no method has been developed that both supports a wide range of models and generates plausible counterfactual sequences.

Therefore, this study proposes an approach with the following three steps. (1) First, the distances between sequential data are calculated using weighted Levenshtein distance, and prototypes (typical patterns) are extracted by clustering using k-medoids and then visualized. (2) Next, using these distances, predictions are made using nonparametric regression techniques such as k-nearest neighbor (k-NN). (3) Finally, based on these predictions, measures are proposed to improve specific sequences using counterfactual explanation. The effectiveness of both steps was evaluated using actual survey data; based on this, a new survey method was proposed for obtaining sequential data.

The rest of the manuscript is organized as follows. Section 2 describes the proposed method. Then, Section 3 describes the experimental details and the results, and Section 4 discusses the results with respect to the aim of the study and presents the limitations. Finally, Section 5 provides the concluding remarks.

\section{Method}

Section 2.1 introduces the problem to be solved in the study, and Section 2.2 describes using the proposed method to solve it.

\subsection{Definition and Problem Setting}

We propose the following definition with reference to customer journey studies such as those of Bernard and Andritsos (2019) and the serial data mining work of Zhang et al.\ (2001) \cite{ref13,ref14,ref27}.

First, let touchpoints be defined as \(\mathcal{TP} = \{ \mathit{tp}_1, \mathit{tp}_2, \ldots, \mathit{tp}_e \}\), where "touchpoints" refers to the means and places (e.g., product reviews, social media, e-commerce sites, in-store) where consumers interact with a product.

Furthermore, let \(\mathcal{A} = \{ a_1, \ldots, a_h \}\) denote \(h\) actions (including no action), such as performing searches, making purchase or non-purchase decisions, and writing product reviews.

Let \(\mathcal{ST} = \langle \mathit{st}_1, \ldots, \mathit{st}_m \rangle\) be a set of \(m\) ordered stages, where "stage" refers to steps in the consumption process, namely the pre-purchase, purchase, and post-purchase stages.

Then, the \(q\)th data point of individual \(i\) is represented as \(X_q^i \in (\mathit{tp}, a; \mathit{st})\), where \(\mathit{tp} \in \mathcal{TP}\), \(a \in \mathcal{A}\), and \(\mathit{st} \in \mathcal{ST}\).

The sequential data belonging to individual \(i\) is denoted by the ordered dataset \(s_i = \langle X_1^i, \ldots, X_{n_i}^i \rangle\), where \(n_i\) is the total number of observations for individual \(i\).

The subsequence in which each pair \((\mathit{tp}, a)\) belongs to the \(g\)th stage \(\mathit{st}_g\) is denoted as \(s_i(\mathit{st}_g)\).

Based on the above definitions, let \(\mathcal{S} = \{ s_1, s_2, \ldots, s_r \}\) be a set of sequences for \(r\) individuals. Let \(\mathcal{S}(\mathit{st}_g)\) be the set of subsequences \(s_i(\mathit{st}_g)\) belonging to stage \(\mathit{st}_g\).

Finally, let \(\mathbf{y} = \{ y_1, y_2, \ldots, y_r \}\) represent the variables to be predicted, where each \(y_i\) can be derived from \((\mathit{tp}, a; \mathit{st})\), conveying information such as whether a purchase is made at the purchase stage.

\textbf{Problem:} Given the sequence set \(\mathcal{S}\), this study aims to (1) extract \(k\) prototypes based on the distances between sequences, and (2) predict the outcome variable \(\mathbf{y}\) using these distances.

\subsection{Proposed Method}
\subsubsection{Definition of the Distances Between Sequences}

This section defines the distance between the sequences of \( S \). The Levenshtein distance (LD), a type of edit distance, is used. The data \( X_q^i \) are not usually interchanged across stages \cite{ref28, ref29}. The importance of each stage should be considered; for example, the presence or absence of purchases can result in large differences. However, conventional LD does not account for such importance, and thus fails to capture these differences appropriately.

To include such a condition, the distance between the two sequences \( s_i \) and \( s_j \) is defined as the weighted sum of the distances at each stage, as follows:

\begin{equation}
D_l(s_i, s_j) = \sum_{g=1}^{m} w_g \, d_l\left(s_i(st_g), s_j(st_g)\right)
\end{equation}

where \( w_g \geq 0 \) is the weight, and \( d_l \) is the Levenshtein distance. These distances are computed by considering each pair \( (tp, a) \) in \( s_i(st_g) \) as a single symbol.

The Levenshtein distance represents the number of operations (insertions, deletions, substitutions) required to transform one string into another. It satisfies properties such as the triangle inequality \cite{ref30}, and the above formulation inherits this property.

\paragraph{Proposition.}
If \( D_l(s_i, s_j) = \sum_{g=1}^{m} w_g \, d_l(s_i(st_g), s_j(st_g)) \), \( w_g \geq 0 \), and each \( d_l(s_i(st_g), s_j(st_g)) \) satisfies the triangle inequality, then \( D_l(s_i, s_j) \) also satisfies the triangle inequality.

\paragraph{Proof.}
To simplify notation, let \( D_l(X, Y) = \sum_{g=1}^{m} w_g \, d_l(x_g, y_g) \), where \( x_g \), \( y_g \) are the \( g \)th components of sequences \( X \), \( Y \). Let

\[
X = (x_1, x_2), \quad Y = (y_1, y_2), \quad Z = (z_1, z_2)
\]

Define:
\begin{align}
D_l(X, Y) &= w_1 d_l(x_1, y_1) + w_2 d_l(x_2, y_2) \\
D_l(X, Z) &= w_1 d_l(x_1, z_1) + w_2 d_l(x_2, z_2) \\
D_l(Y, Z) &= w_1 d_l(y_1, z_1) + w_2 d_l(y_2, z_2)
\end{align}

Assuming that \( d_l \) satisfies the triangle inequality, we have:
\begin{align}
d_l(x_1, z_1) &\leq d_l(x_1, y_1) + d_l(y_1, z_1) \label{ineq1} \\
d_l(x_2, z_2) &\leq d_l(x_2, y_2) + d_l(y_2, z_2) \label{ineq2}
\end{align}

Multiplying \eqref{ineq2} by \( w_2 \geq 0 \):
\begin{equation}
w_2 d_l(x_2, z_2) \leq w_2 d_l(x_2, y_2) + w_2 d_l(y_2, z_2) \label{ineq3}
\end{equation}

Adding \( w_1 d_l(x_1, z_1) \) to both sides of \eqref{ineq3}:
\begin{equation}
w_1 d_l(x_1, z_1) + w_2 d_l(x_2, z_2) \leq w_1 d_l(x_1, z_1) + w_2 d_l(x_2, y_2) + w_2 d_l(y_2, z_2) \label{ineq4}
\end{equation}

From \eqref{ineq1}, we have:
\begin{equation}
w_1 d_l(x_1, z_1) \leq w_1 d_l(x_1, y_1) + w_1 d_l(y_1, z_1) \label{ineq5}
\end{equation}

Substituting \eqref{ineq5} into \eqref{ineq4}, we get:
\begin{align*}
w_1 d_l(x_1, z_1) + w_2 d_l(x_2, y_2) + w_2 d_l(y_2, z_2) &\leq w_1 d_l(x_1, y_1) + w_1 d_l(y_1, z_1) \\
&\quad + w_2 d_l(x_2, y_2) + w_2 d_l(y_2, z_2)
\end{align*}

Hence,
\[
D_l(X, Z) \leq D_l(X, Y) + D_l(Y, Z)
\]

This conclusion also holds for the general case, where
\[
D_l(X, Y) = \sum_{g=1}^{m} w_g \, d(x_g, y_g)
\quad \Longleftrightarrow \quad
D_l(s_i, s_j) = \sum_{g=1}^{m} w_g \, d_l\bigl(s_i(st_g), s_j(st_g)\bigr).
\]

Therefore, \( D_l(s_i, s_j) \) satisfies the triangle inequality and thus defines a valid distance metric over the space of staged sequences.
\hfill \ensuremath{\square}

\paragraph{}
The costs of replacement, insertion, and deletion operations can be modified. However, using LD with arbitrary weights or normalization may violate the distance axioms. Therefore, we adopt the above weighted LD formulation, which respects distance properties. Notably, the total computation cost is reduced by calculating distances separately for each stage \( st_g \).

\subsubsection{Prototype Detection and Visualization}

This section describes how the proposed weighted Levenshtein distance is used to extract prototypes. K-medoids, a clustering method, is employed for this purpose \cite{ref31}. K-medoids identifies data points, called medoids, that minimize the total dissimilarity to all other data in the same cluster. These medoids serve as the prototypes. Since only pairwise distances between sequences are available, this method enables prototype detection without requiring data in vector format, which is commonly assumed in conventional clustering methods.

Other methods such as K-means are not suitable in this context, as they require the computation of mean vectors, which is not possible when only pairwise sequence distances are provided.

\paragraph{Visualization.}
Let \( D \) be a matrix whose entries represent the pairwise distances \( D_l \). Each sequence is visualized as a point in a two-dimensional space based on the concept of multidimensional scaling (MDS) \cite{ref32}. Although other dimensionality reduction techniques such as principal component analysis (PCA) exist, MDS is appropriate here because it requires only a distance matrix and not feature vectors.

Let \( I \) be an identity matrix and \( \mathbf{1}_r \) an \( r \)-dimensional column vector of ones. The procedure for MDS-based visualization is as follows:

\begin{enumerate}
\item \textbf{Represent the centering matrix:}
\begin{equation}
D_{ce} = -\frac{1}{2} C D^2 C, \quad \text{where} \quad C = I - \frac{1}{r} \mathbf{1}_r \mathbf{1}_r^\prime
\end{equation}

\item \textbf{Perform eigenvalue decomposition of \( D_{ce} \):}
\begin{equation}
D_{ce} = V \Lambda V^{-1}
\end{equation}
where \( \Lambda = \begin{bmatrix}
\lambda_1 & 0 & 0 \\
0 & \lambda_2 & 0 \\
0 & 0 & \ddots
\end{bmatrix} \) contains the eigenvalues and \( V = [\vec{v}_1, \vec{v}_2, \ldots] \) is the matrix of eigenvectors.

\item \textbf{Compute the coordinate matrix:}
\[
X^* = V \Lambda^{1/2}
\]
assuming only the top two eigenvalues \( \lambda_1 \) and \( \lambda_2 \) are nonzero.

\item \textbf{Place each data point using \( X^* \).}
\end{enumerate}

The resulting two-dimensional coordinates represent each sequence, and the Euclidean distances between the points approximate the original sequence distances. This visualization supports interpretation of the cluster structure and the distribution of prototype sequences.

\subsubsection{Prediction and Counterfactual Explanation}

This section explains how prediction was performed using the proposed Levenshtein distance. In this study, the $k$-nearest neighbors ($k$-NN) method was employed. This is a nonparametric regression approach that leverages distance, and its use here is consistent with earlier sections: since only pairwise distances between sequences and corresponding target values $y$ are available, conventional regression or machine learning models that require feature vectors cannot be applied.

To avoid confusion with $k$ in $k$-medoids clustering, we use $k'$ to denote the number of neighbors in $k$-NN.

The prediction for a given sequence $s_i$ is computed as:
\begin{equation}
\hat{y}(s_i) = \frac{1}{k'} \sum_{s_j \in N_i} y_j
\end{equation}
where $N_i$ is the set of $k'$ nearest sequences to $s_i$ under the proposed distance measure.

\paragraph{Counterfactual Explanation.}
Additionally, a new type of counterfactual explanation is proposed as an interpretability method in machine learning \cite{ref23, ref24}. For example, this method suggests an alternative customer journey that changes a non-purchase prediction into a purchase outcome, thereby highlighting factors that influence consumer decisions.

Specifically, the counterfactual sequence $\text{cfs}_i$ for a given $s_i$ is defined as:
\begin{equation}
\text{cfs}_i = \arg\min_{s_j \in S \setminus \{s_i\}} \left\{ \mathrm{loss}(y_{\mathrm{obj}}, y_j) + \lambda \cdot D'(s_i, s_j) \right\}
\end{equation}
where:
\begin{itemize}
    \item $S \setminus \{s_i\}$ is the set of all sequences excluding $s_i$,
    \item $y_{\mathrm{obj}}$ is the desired prediction outcome (e.g., purchase),
    \item $\lambda$ is a hyperparameter that balances prediction loss and similarity,
    \item $D'$ is a potentially modified distance between sequences (possibly restricted to some stages $st_g$).
\end{itemize}

A novel aspect of this method is that it does not generate new (synthetic) sequences; rather, it selects the most plausible counterfactual sequence from the existing data. This contributes to producing more realistic and actionable explanations, which has not been widely addressed in prior work.

The core idea is to detect an existing sequence that achieves the desired target $y_{\mathrm{obj}}$ and is also close to $s_i$ under the defined distance metric. In practice, the process computes the objective value for all candidate sequences and selects the one with the smallest score. The change from the original $s_i$ to the selected counterfactual is analyzed to identify which components (e.g., touchpoints or actions) differ. Based on this insight, specific suggestions can be made to improve customer journeys.

\section{Experimental}

Section 3.1 describes the data used in the analysis, and Section 3.2 discusses the results.

\subsection{Data}

Although automatically collected log data from e-commerce sites can be used to analyze customer journeys, acquiring such data for general product purchases presents practical challenges. Therefore, we developed and employed a new questionnaire-based method that allowed the number of recorded actions to vary by individual.

The survey was conducted online and focused on the most recently purchased type of cosmetics—eye shadows. It was administered to 127 female university students in Japan over a two-day period from November 30 to December 1, 2022. Participants who indicated an interest in eye shadow were first asked: “How did you come to know about eye shadow?” Subsequent questions such as “Which of the following did you do next?” were then presented, with responses recorded up to 10 times per participant.

The questionnaire consisted of 13 predefined combinations of touchpoints and actions:

\begin{enumerate}[label=\alph*.]
\item Knowing the product through TV commercials [awareness]
\item Knowing the product through direct word-of-mouth communication [awareness]
\item Understanding the product through word-of-mouth communication on social media [awareness]
\item Knowing the product through other methods [awareness]
\item Comparison and confirmation through various websites by searching [information gathering]
\item Comparison and confirmation through e-commerce sites [information gathering]
\item Comparison and confirmation through stores [information gathering]
\item Comparison and confirmation through social media [information gathering]
\item Performing purchases on e-commerce sites [purchase]
\item Performing purchases in stores [purchase]
\item Not performing purchases [non-purchase]
\item Writing product reviews on social media [post-purchase]
\item Other
\end{enumerate}

Hereafter, each symbol corresponds to the meanings described below and is also consistent with the formal representation introduced in Section~2.1. 
The symbols $a$ to $h$ represent the subset of combinations of $(tp, a)$, where $tp \in \{\text{TV}, \text{person}, \text{social media}, \text{e-commerce site}, \text{storefront}\}$ and $a \in \{\text{cognition}, \text{information gathering}\}$ (corresponding to the options described above). 
Here, $i$ and $j$ correspond to $(tp = \text{"e-commerce site"}, a = \text{"purchase"})$ and $(tp = \text{"storefront"}, a = \text{"purchase"})$, respectively. 
Finally, purchases in $k$ are represented by ``1'' and non-purchases by ``0''.

The sequences were classified into three stages: \( st_1 \) (pre-purchase: first recognition), \( st_2 \) (pre-purchase: awareness and information gathering), and \( st_3 \) (purchase stage: purchase and non-purchase). Post-purchase actions (such as items \textit{l} and \textit{m}) were excluded from the analysis.

Sequences that did not include either a purchase or non-purchase action were omitted, as were sequences with logically inconsistent stage transitions (e.g., taking a pre-purchase action after a purchase action). Only valid paths were retained, such as: \( st_1 \rightarrow st_1 \) (multiple awareness), \( st_1 \rightarrow st_2 \) (consideration after awareness), \( st_2 \rightarrow st_2 \) (multiple considerations), \( st_2 \rightarrow st_1 \) (awareness after consideration), \( st_1 \rightarrow st_3 \) (purchase after awareness), and \( st_2 \rightarrow st_3 \) (purchase after consideration).

As a result, 104 valid samples were obtained. Of these, 86 resulted in a purchase and 18 in a non-purchase.

To facilitate the recall of recent consumption behavior, respondents were first asked to describe specific brand names and shopping contexts. However, this approach may introduce recall bias—a limitation that is acknowledged in the design of survey-based research.

\subsection{Results}

Section 3.2.1 presents descriptive statistics, Section 3.2.2 discusses the results of prototype detection and visualization, and Section 3.2.3 reports the results of prediction and counterfactual explanation.

\subsubsection{Descriptive Statistics}

First, the descriptive statistics for each stage of the obtained data are presented.

For the first stage, \( st_1 \), that is, the pre-purchase (first recognition) phase, the results are shown in Table~\ref{tab:stage1}.

\begin{table}[htbp]
\centering
\caption{Descriptive statistics of items in Stage 1}
\includegraphics[width=0.35\linewidth]{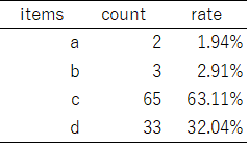}
\label{tab:stage1}
\end{table}

The results show that item~\textit{c}, “Understanding the product through word-of-mouth communication on social media,” appears most frequently, followed by item~\textit{d}, “Knowing the product through other methods.” This indicates that recognition through social media is the most common. Item~\textit{d} is considered to include recognition through offline sources such as in-store exposure.

For the next stage, \( st_2 \), the pre-purchase phase (awareness and information gathering), the results are shown in Table~\ref{tab:stage2_1} and Table~\ref{tab:stage2_2}.

\begin{table}[htbp]
\centering
\caption{Descriptive statistics of items in Stage 2}
\includegraphics[width=0.35\linewidth]{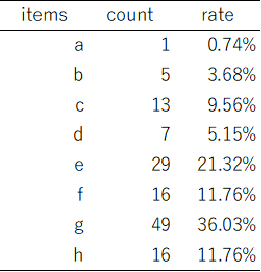}
\label{tab:stage2_1}
\end{table}

\begin{table}[htbp]
\centering
\caption{Per-sample item count statistics}
\includegraphics[width=0.35\linewidth]{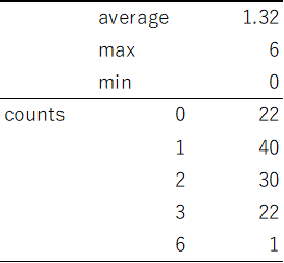}
\label{tab:stage2_2}
\end{table}

The results show that item~\textit{g}, “Comparison and confirmation through stores,” was the most frequent, followed by item~\textit{e}, “Comparison and confirmation through various websites by searching,” and item~\textit{h}, “Comparison and confirmation through social media.” In this stage, the number of actions ranged from a minimum of 0 to a maximum of 6, with an average of 1.32 items. In particular, it was revealed that individuals typically performed 1 to 3 actions during this phase. On the other hand, a value of 0 indicates that the purchase was made immediately after recognition, suggesting behavior similar to impulse buying. Taken together, these results suggest that after recognizing a product, consumers generally tend to compare or confirm it in physical stores.

For the final stage, \( st_3 \), the number of purchases and non-purchases was 86 and 18, respectively. This indicates that most respondents chose to make a purchase.

\begin{figure}[htbp]
\centering
\includegraphics[width=0.7\linewidth]{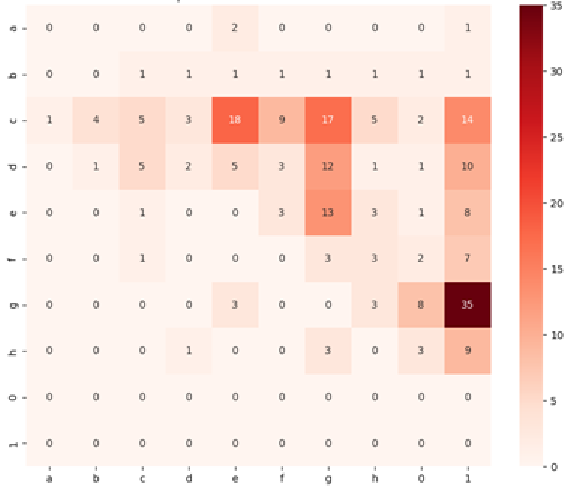}
\caption{Sequential Co-occurrence Matrix (Y-axis: From Item, X-axis: To Item)}
\label{fig:cooccur}
\end{figure}

Finally, all stages were integrated, and a matrix of item pairs was created with consideration of the sequence order. The result was visualized, as shown in Figure~\ref{fig:cooccur}. For example, for a sequence such as \textit{a, e, i}, the transitions \( a \rightarrow e \) and \( e \rightarrow i \) were each counted. Only item pairs that satisfy the stage transition constraints described in Section~3.1 (e.g., from \( st_1 \) to \( st_2 \), or from \( st_2 \) to \( st_3 \)) were used.

The results show that in transitions from Stage \( st_1 \) (a, b, c, d) to the next stage, the most frequent item pairs were \( c \rightarrow e \), \( c \rightarrow g \), \( c \rightarrow i \), and \( d \rightarrow g \). This suggests that after recognizing a product through social media, consumers often proceed to consider it through websites or in physical stores. In transitions from Stage \( st_2 \) (e, f, g, h) to the next stage, \( g \rightarrow i \) was by far the most frequent, followed by \( e \rightarrow g \) and \( h \rightarrow i \). This indicates that many consumers make a purchase after considering the product in stores.

However, it should be noted that these results are based solely on pairwise frequency. To investigate customer journeys involving more than two consecutive actions, full sequence analysis is necessary.

\subsubsection{Results of Prototype Detection and Visualization}

The analysis of the results is presented here. First, the clustering results are described. Four different distance measures were considered: the standard Levenshtein distance (LD), the Damerau-Levenshtein distance, and two weighted versions with weights \( w_1 = 2, w_2 = 1, w_3 = 1 \) and \( w_1 = 2, w_2 = 1, w_3 = 10 \), respectively. Clustering was evaluated using the Silhouette Coefficient (SC) for cluster sizes \( k \in \{2, 3, 4, 5, 6, 7, 8\} \) \cite{ref33}. Initial values were set using the k-medoids++ algorithm, and random seeds were fixed to ensure reproducibility. Higher SC values indicate better clustering performance. In general, SC can be high for smaller \( k \), but higher values may be preferred depending on the application.

\begin{table}[htbp]
\centering
\caption{Clustering results (Silhouette Coefficient for different settings)}
\includegraphics[width=0.9\linewidth]{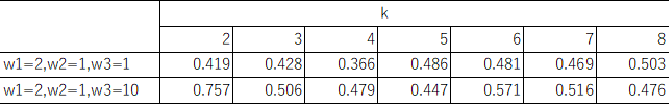}
\label{tab:clustering}
\end{table}

Table~\ref{tab:clustering} shows that SC values increased when higher weights were assigned to \( w_3 \). This reflects the greater influence of differences in the purchase stage \( st_3 \) under such settings. The typical sequences extracted under the setting \( w_1 = 2, w_2 = 1, w_3 = 10 \) with \( k = 6 \) are shown below:

\begin{enumerate}
\item [1:] \texttt{[c, e, g, 1]}
\item [2:] \texttt{[c, e, g, 0]}
\item [3:] \texttt{[c, g, 1]}
\item [4:] \texttt{[d, c, 1]}
\item [5:] \texttt{[c, b, e, g, 1]}
\item [6:] \texttt{[d, 1]}
\end{enumerate}

The cluster sizes were as follows: Cluster~1: 14, Cluster~2: 18, Cluster~3: 35, Cluster~4: 9, Cluster~5: 7, and Cluster~6: 20. The most common pattern was Pattern~3, which represented customers who recognized a product through social media, checked it in stores, and then purchased it. The next most common was Pattern~6, which represented customers who purchased a product after recognizing it through other means—possibly even prior familiarity. Pattern~2 and Pattern~5 represent customers who performed online searches, confirmed products in stores, and either made a purchase or did not, depending on the situation.

\begin{figure}[htbp]
\centering
\includegraphics[width=0.7\linewidth]{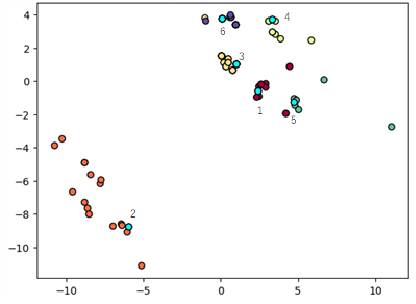}
\caption{Visualization of the sequence clusters in 2D space}
\label{fig:cluster_map}
\end{figure}

Next, the visualization results are presented in Figure~\ref{fig:cluster_map}. Although the positions are relative, patterns like 1 and 5 appeared in the lower right, suggesting more careful judgment. Conversely, patterns like 3 and 6 occupied more central or upper positions, indicating more direct decision-making. The non-purchase group also tended to cluster in areas suggesting more deliberative behavior.

\subsubsection{Results of Prediction and Counterfactual Explanation}

Then, the $k$-NN algorithm was used to predict $y$. Specifically, in this study, \( S(st_1) \) and \( S(st_2) \) were used as explanatory variables, and the purchase (“1”) or non-purchase (“0”) decision in \( S(st_3) \) was used as the target variable $y$. Regarding the data, 80\% and 20\% were used as training and test data, respectively. The prediction was performed using \( k' \in \{1,2,3,4,5\} \), and each setting was repeated 100 times to evaluate the mean and variance of both accuracy and F1-measure. These results are shown in Table~\ref{tab:knn}.

\begin{table}[htbp]
\centering
\caption{Comparison of accuracy and F1-measure values}
\includegraphics[width=0.5\linewidth]{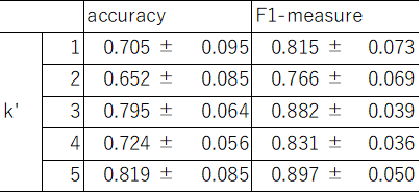}
\label{tab:knn}
\end{table}

Comparing the values of each model, the accuracy for \( k' = 5 \) was approximately 0.81, indicating that it would be possible to predict whether a product will be purchased based on the sequence.

The results of the counterfactual explanation for the $k$-NN model using sequential data are presented below. Specifically, four cases are shown in which the original sequence had a label of $y = 0$ (non-purchase), and an alternative sequence improved the prediction to \( \hat{y} = 1 \) (purchase).

\vspace{0.5em}
\noindent
(Base sequence and counterfactual sequence: 1)\\
Base sequence: \texttt{["c", "c", "e", "g"]} :\quad $y=0$\\
Counterfactual sequence: \texttt{["c", "b", "e", "g"]}: \quad $\hat{y}=1$\\
The improved sequence was derived by replacing \texttt{c}: “Knowing about the product through word-of-mouth communication on social media” with \texttt{b}: “Knowing about the product through direct word-of-mouth communication,” indicating that the respondent might purchase the product. This suggested the importance of direct word-of-mouth communication over communication on social media with respect to purchasing decisions.

\vspace{0.5em}
\noindent
(Base sequence and counterfactual sequence: 2)\\
Base sequence: \texttt{["d", "e", "f"]} :\quad $y=0$\\
Counterfactual sequence: \texttt{["c", "e", "f"]}: \quad $\hat{y}=1$\\
The improved sequence was derived by replacing \texttt{d}: “Knowing the product through other methods” with \texttt{c}: “Knowing the product through word-of-mouth communication on social media,” indicating that the respondent might purchase the product. This suggested the importance of social media communication over more ambiguous or less direct information sources.

\vspace{0.5em}
\noindent
(Base sequence and counterfactual sequence: 3)\\
Base sequence: \texttt{["d", "g", "h"]} :\quad $y=0$\\
Counterfactual sequence: \texttt{["d", "g", "e"]}: \quad $\hat{y}=1$\\
The improved sequence was derived by replacing \texttt{h}: “Comparison and confirmation through social media” with \texttt{e}: “Comparison and confirmation through various websites by searching,” indicating that the respondent might purchase the product. This suggested that broader web-based comparison may be more effective than relying solely on social media content.

\vspace{0.5em}
\noindent
(Base sequence and counterfactual sequence: 4)\\
Base sequence: \texttt{["b"]}:\quad $y=0$\\
Counterfactual sequence: \texttt{["c"]}: \quad $\hat{y}=1$\\
The improved sequence was derived by replacing \texttt{b}: “Knowing the product through direct word-of-mouth communication” with \texttt{c}: “Understanding the product through word-of-mouth communication on social media,” indicating that the respondent might purchase the product. This suggested that communication on social media might have a broader or more persuasive influence compared to direct word-of-mouth interactions.

\vspace{1em}
This analysis demonstrated that even minor changes in the sequence of information acquisition and evaluation can influence purchase intentions. The following key findings emerged:

\begin{enumerate}
\item The mode of word-of-mouth communication (direct vs.\ social media-based) significantly impacts purchasing decisions, with effectiveness varying by context.
\item Social media-based information tends to be more persuasive than impersonal or unspecified sources.
\item During the comparison and confirmation stage, web searches offering diverse information may be more effective than social media-based content.
\end{enumerate}

These findings suggest that strategically designing information channels is essential for encouraging consumer purchases. Moreover, these results clearly showed that it was possible to propose ways to change a particular preselected sequence such that it would lead to a purchase. Furthermore, they indicated the factors that were important for performing purchases in each sequence.

\section{Discussion}

This study was conducted to formally and quantitatively analyze customer journeys. The differences between previous studies and the proposed study are discussed below.

Unlike studies that use mixed Markov models, our study specifically detects typical cases~\cite{ref12}. Then, there are several studies that have applied $k$-medoids to edit distances such as LD~\cite{ref13}. Our study differs from those in that it defines a new distance for more general customer journey data with categories such as stages. Since this distance satisfies the axiom of distance, it can easily be extended to other methods.

In addition, this study proposed a visualization method using MDS based on the proposed distance. Unlike previous research, the distance used here is not a simple Levenshtein distance but is designed for marketing-specific applications. This approach not only enables the extraction of prototypes, but also allows for identifying their relative positioning, which facilitates interpretation as prototypes and evaluation of their validity.

Furthermore, our study proposes not only clustering but also a method to predict and derive counterfactual hypothetical cases. This approach also contributes to the research field of interpretable counterfactual explanations—or algorithmic recourse—from the following perspectives.

First, (i) \textbf{plausibility}: Each suggested sequence corresponds to an actual customer journey observed in the dataset. Counterfactual recommendations are generated based on sequences that were empirically observed.

Second, the method improves (ii) \textbf{managerial interpretability}: Since each edit operation can be directly associated with a specific marketing intervention (e.g., replacing the first touchpoint from “direct word-of-mouth” to “social media-based word-of-mouth”). Recent studies have shown that the usefulness of counterfactual explanations heavily depends on such validity.

Finally, it introduces (iii) \textbf{a novel approach for sequential data}: While robustness in structured data has been extensively studied, for sequential data, only recently have methods begun to address symbolic constraints on allowable edits. This study offers one such method.

However, establishing robustness guarantees for the proposed approach remains an important topic for future work.

Moreover, we propose a new survey method and a method for cleansing data rather than just providing a formal definition of data. These two are novelties and contributions in this study.

Regarding the shortcomings of the proposed study, they are presented below.

\begin{enumerate}
    \item First, a customer journey not only includes different stages but also the actual behaviors (search, purchase, and non-purchase) and psychological states of consumers (such as perceived needs) and more specific information at touchpoints~\cite{ref10,ref15}. Therefore, it is necessary to incorporate and extend this information. Since these pieces of information may have complex structures, distances between sequences can be obtained by using approaches like tree structures and graphs.
    
    \item Another issue is estimating the level of awareness of consumers regarding their behavior and changes in their awareness.

    \item Finally, regarding problems such as recall bias in data collection, a method that enables automatic data collection should be considered.
\end{enumerate}

\section{Conclusions}

The overall contributions of the study are listed below.

\begin{enumerate}
    \item A method to formally define data sequences and the distances between them using Levenshtein distance (LD) was proposed.
    \item Typical cases were extracted using $k$-medoids. Furthermore, a method for visualizing these data was presented.
    \item Both a method for predicting purchases using $k$-NN with the proposed distance and a method for extracting counterfactual explanations to interpret the prediction results were proposed.
    \item A new survey method to obtain the data used in the study was proposed.
\end{enumerate}

The survey-based data analysis revealed typical interpretable patterns. Furthermore, through visualization, the positioning of the data could be understood, predictions could be made regarding whether purchases would be performed, and counterfactual hypothetical cases and important elements could be detected.

The analysis and research methods used in this study will be useful for understanding the future behavior of consumers and designing consumer experiences~\cite{ref7}.

\section*{Acknowledgments}

I would like to thank Yui Nakazawa (Faculty of Business, Kyoritsu Women's University) for her help in designing and collecting the data used in the study. This study was supported by a JSPS Grant-in-Aid for Scientific Research (25K05381).

\bibliographystyle{plain}
\bibliography{references}

\end{document}